\pdfoutput=1

\documentclass[11pt]{article}

\usepackage{ACL2023}

\usepackage{times}
\usepackage{latexsym}

\usepackage[T1]{fontenc}

\usepackage[utf8]{inputenc}

\usepackage{microtype}

\usepackage{inconsolata}

\usepackage{xcolor}
\usepackage{amssymb}

\definecolor{leacolor}{rgb}{0.40,0.60,1.00}
\definecolor{selcolor}{rgb}{0.53,0.30,1.00}
\definecolor{mmcolor}{rgb}{0.1,0.2,0.4}
\definecolor{urjacolor}{rgb}{0.039, 0.478, 0.741}

\definecolor{todocolor}{rgb}{1.00,0.75,0.00}
\definecolor{changedcolor}{rgb}{0.42,0.27,0.57}
\definecolor{removedcolor}{rgb}{0.867,0.176,0.361}

\usepackage{listings}
\usepackage{xcolor}
\colorlet{punct}{red!60!black}
\definecolor{background}{HTML}{EEEEEE}
\definecolor{delim}{RGB}{20,105,176}
\lstdefinelanguage{json}{
    basicstyle=\normalfont\ttfamily,
    numbers=left,
    numberstyle=\tiny,
    stepnumber=0,
    numbersep=1pt,
    showstringspaces=false,
    breaklines=true,
    frame=lines,
    backgroundcolor=\color{background},
    literate=
      {:}{{{\color{punct}{:}}}}{1}
      {,}{{{\color{punct}{,}}}}{1}
      {\{}{{{\color{delim}{\{}}}}{1}
      {\}}{{{\color{delim}{\}}}}}{1}
      {[}{{{\color{delim}{[}}}}{1}
      {]}{{{\color{delim}{]}}}}{1},
}
\usepackage{booktabs}
\usepackage{graphicx}
\usepackage{listings}
\usepackage{multirow}

\usepackage[inline]{enumitem}

\usepackage[framemethod=TikZ]{mdframed}
\newcounter{theo}[section] 
\renewcommand{\thetheo}{\arabic{section}.\arabic{theo}}
\newenvironment{theo}[2][]{%
\refstepcounter{theo}%
\ifstrempty{#1}%
{\mdfsetup{%
frametitle={%
\tikz[baseline=(current bounding box.east),outer sep=0pt]
\node[anchor=east,rectangle,fill=darkblue!20]
{\strut Prompt~\thetheo};}}
}%
{\mdfsetup{%
frametitle={%
\tikz[baseline=(current bounding box.east),outer sep=0pt]
\node[anchor=east,rectangle,fill=darkblue!20]
{\strut Prompt~\thetheo:~#1};}}%
}%
\mdfsetup{innertopmargin=10pt,linecolor=darkblue!20,%
linewidth=2pt,topline=true,font=\small,%
frametitleaboveskip=\dimexpr-\ht\strutbox\relax,
backgroundcolor=lightgray!20
}
\begin{mdframed}[nobreak=true]\relax%
\label{#2}}{\end{mdframed}}

\title{Leveraging Few-Shot Data Augmentation and Waterfall Prompting for Response Generation} 

\author{
  Lea Krause \\
  Vrije Universiteit Amsterdam \\
  \texttt{l.krause@vu.nl} \\ \And
  Selene Báez Santamaría \\ \
  Vrije Universiteit Amsterdam \\
  \texttt{s.baezsantamaria@vu.nl} \\ \AND
  Michiel van der Meer \\ Leiden University \\
  \texttt{m.t.van.der.meer@liacs.leidenuniv.nl} \\ \And
  Urja Khurana \\
  Vrije Universiteit Amsterdam \\
  \texttt{u.khurana@vu.nl}\\
}

\begin{document}
\maketitle
\begin{abstract}
This paper discusses our approaches for task-oriented conversational modelling using subjective knowledge, with a particular emphasis on response generation.
Our methodology was shaped by an extensive data analysis that evaluated key factors such as response length, sentiment, and dialogue acts present in the provided dataset. We used few-shot learning to augment the data with newly generated subjective knowledge items and present three approaches for DSTC11: (1) task-specific model exploration, (2) incorporation of the most frequent question into all generated responses, and (3) a waterfall prompting technique using a combination of both GPT-3 and ChatGPT.
\end{abstract}

\section{Introduction}
Task-Oriented Dialogue (TOD) Systems are traditionally designed to facilitate users in achieving specific objectives, such as looking up train times or booking a flight in a dialogue setting. For these tasks, the models are often given access to a database of factual information to complete the task. However, other tasks necessitate not only factual but also subjective insights, which are derived from other users' opinions. Handling subjective knowledge and using it for generating dialogue responses is the core of the \textit{Subjective-Knowledge-based Task-Oriented Dialogue (SK-TOD)} \cite{zhao2023others} challenge. The challenge is set up as conversations between users and artificial assistants, inquiring about and potentially booking hotels or restaurants. The organisers provided dialogue snapshots and a knowledge base with subjective reviews and FAQs related to said hotels and restaurants.

\begin{figure}[htp]
    \centering
    \includegraphics[width=\columnwidth]{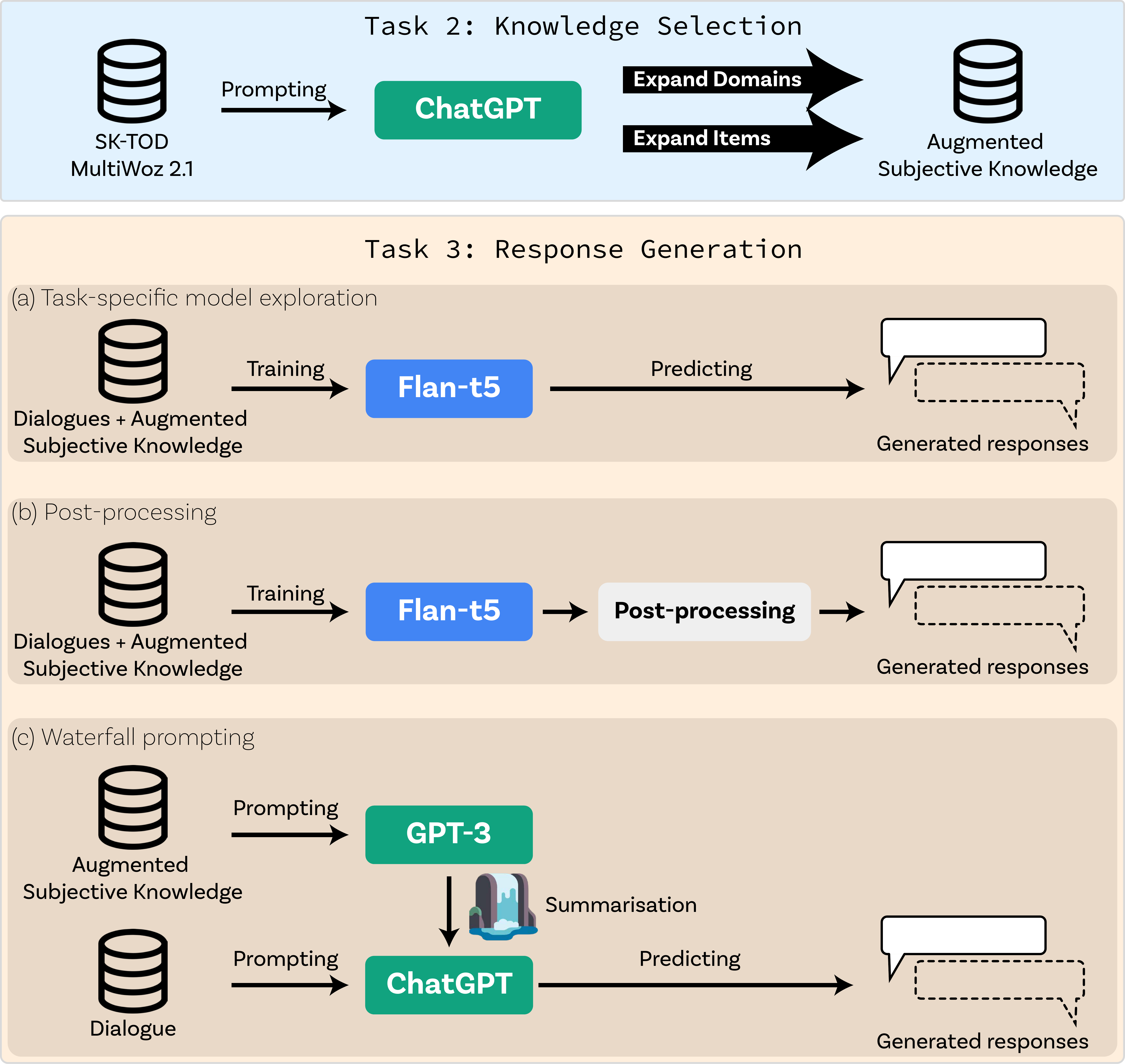}
    \caption{Overview of the approaches for \textit{Knowledge Selection} and \textit{Response Generation}. For approach 3a only the best-performing model is depicted.}
    \label{fig:approaches}
\end{figure}

The challenge consists of three interlinked sub-tasks: 1) \textit{Knowledge Seeking Turn Detection}, where it is determined whether a turn needs knowledge to create an appropriate response; 2) \textit{Knowledge Selection}, where relevant items are selected from the knowledge base; and 3) \textit{Knowledge-Grounded Response Generation}, where the dialogue history and the selected knowledge items must be aggregated to generate a concise response for the user.

Our primary interest was in sub-task 3, but due to the high correlation of performance with sub-task 2, we simultaneously worked on improving the knowledge selection (see Figure \ref{fig:approaches}). Furthermore, we experimented with splitting sub-task 3 into two smaller tasks. First, we perform few-shot data augmentation, supported by a detailed data analysis, to substantially increase the knowledge base and address generalisation to new domains. Despite its marginal impact on the final test set, given no new domains were added, it forms a robust strategy for future generalisations. Second, we undertook the exploration of task-specific models, we found that using a larger model leads to slight improvements in task performance. In addition, we augmented the generated responses by integrating the most frequently asked question, aiming to enhance conversation coherence and engagement.
Lastly, we also employed a waterfall prompting technique that integrated a combination of Large Language Models. This strategy, though it scored lower in quantitative metrics compared to the baseline, possibly due to the abstractive nature of these models, is a promising avenue for further research\footnote{Our code is available at \url{https://github.com/lkra/dstc11-track5/tree/main/CLTeamL}}.

\section{Related work}
\paragraph{Data Augmentation}
In low-resource data scenarios, synthetic data augmentation has been shown to be a cost-efficient and effective way to increase dataset size~\cite{Anaby-Tavor_etal_2020}. Traditional data augmentation techniques for natural language processing often include simple transformations like synonym substitution, random insertion, deletion, and swapping of words~\cite{wei-zou-2019-eda}. A more complex approach is paraphrasing, where the meaning stays the same but words and syntax change, a popular example of this is back-translation in machine translation~\cite{sennrich-etal-2016-improving}. Data augmentation was also used in successful approaches of last year's edition of DSTC~\cite{tian2021tod-da, thulkeDSTC2021}

Recently, the emergence of large language models (LLMs) like GPT-3 \cite{brown2020language} has opened up new avenues for data augmentation. One approach is to prompt these models to generate additional data that fits the distribution of the original dataset \cite{kumar-etal-2020-data, xia-etal-2020-composed, lee2021neural}. This leverages the capacity of LLMs to generate coherent and contextually appropriate sentences, thereby creating augmented data that closely mirrors real-world linguistic diversity.

\paragraph{Waterfall Prompting}
Combining prompts or rerunning the same prompt has shown increased performance and robustness. For example, \citet{singhal2023expertlevel} coin the term \textit{ensemble refinement} which involves using a two-step prompting approach. In the first iteration, multiple answers are generated for a prompt, and in the second iteration, these answers are incorporated into the prompt to develop a final solution that is more robust for the given task. Similar efforts have been explored such as self-consistency \cite{wang2023selfconsistency}, recitation-augmentation~\cite{sun2023recitationaugmented}, and rational-augmentation~\cite{wang2022rationale}. \citet{pitis2023boosted} expand on this framework by incorporating multiple rounds of prompting, each building upon the improvements made in the previous round. However, these still focus on prompting a single model and differ from a typical ensemble where different models' output is aggregated \cite{wang2023selfconsistency}. To the best of our knowledge, there is limited work on waterfall prompting or assembling the responses from different language models.

\paragraph{Previous DSTC editions}
According to system reports from the DSTC9 and DSTC10 challenges, enhancing the performance of sub-task 2 significantly affects the overall task performance~\cite{kim-etal-2020-beyond}. In the DSTC9 iteration, it was found that ensemble model approaches are effective for knowledge selection~\cite{kim-etal-2020-beyond}. In DSTC10, successful approaches used a separate entity tracking component for knowledge selection to narrow down the search space before document ranking~\cite{kim2022knowledge}.

\section{Data Analysis}
\label{sec:data_analysis}
Before developing our approach to the DSTC tasks, we analyse the available data. We \begin{enumerate*}
    \item inspect the correlation between features,
    \item investigate the sentiment of the available data, and
    \item examine the distribution of dialogue act types.
\end{enumerate*}

\begin{figure}[htb]
    \centering
    \includegraphics[width=.5\textwidth]{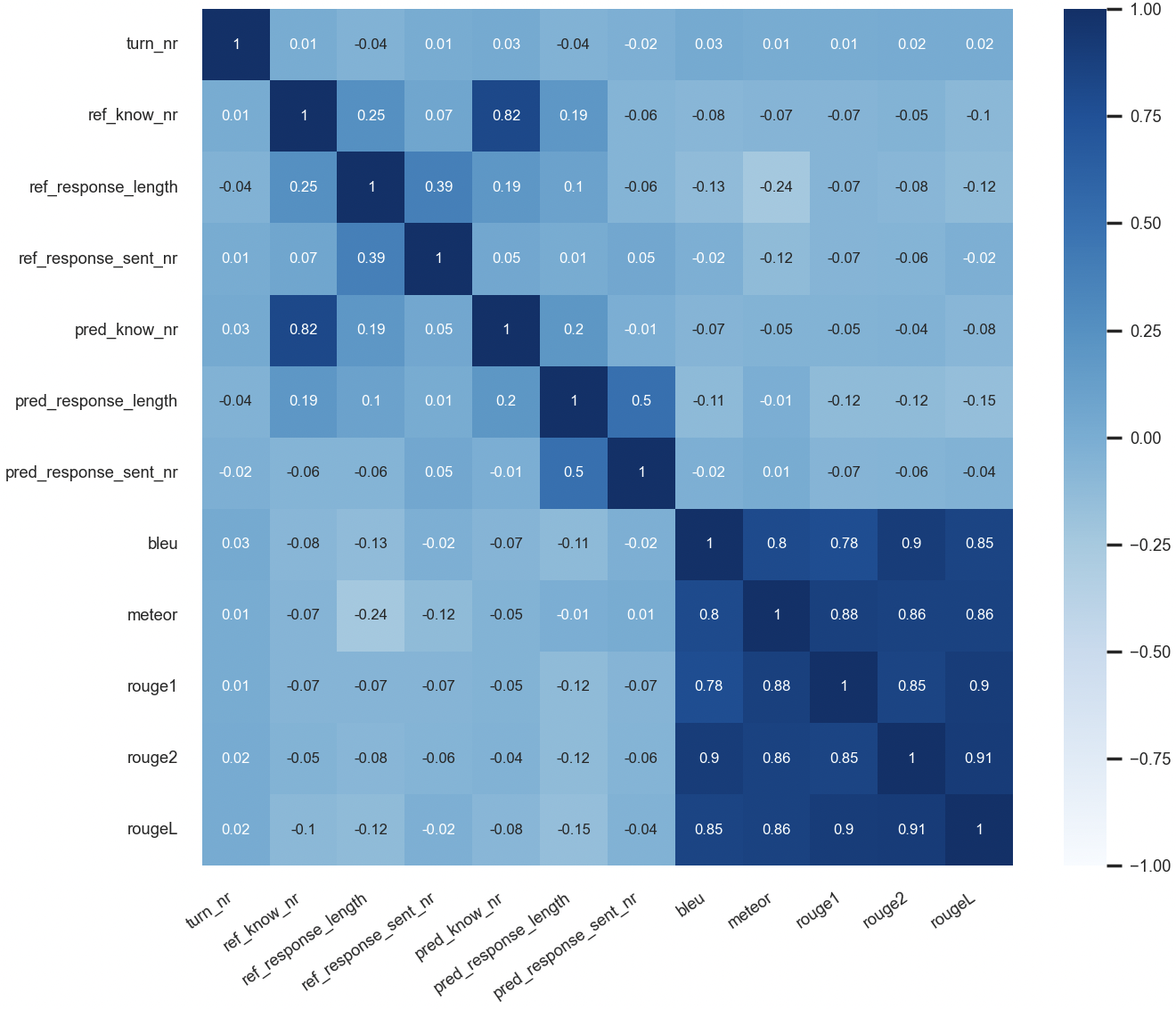}
    \caption{Correlation between reference data, baseline predictions, and automatic scores, on the validation set.}
    \label{fig:correlations_baseline}
\end{figure}

\subsection{Feature correlation}
To steer our approach, we investigate the potential influence of various factors on automatic metric scores. We examine correlations between the following variables: \begin{enumerate*}[label=(\arabic*)]
    \item the number of dialogue turns,
    \item the number of selected knowledge items, and
    \item the lengths of both the reference and prediction in terms of characters and sentences
\end{enumerate*} (Figure \ref{fig:correlations_baseline}).

\noindent After not finding strong correlations among these parameters, we decide to split the response into two sub-tasks. Namely, we process system responses by separating them into two parts: \begin{verbatim}
   summary + [optional] question
\end{verbatim}

These two parts are defined as follows:
\begin{itemize}
    \item Summary: This part of the response focuses on summarising the selected knowledge items.
    \item Question: If included, this part of the response pertains to dialogue management, where the system can offer assistance in continuing with the current task, such as making a reservation.
\end{itemize}

\subsection{Summary analysis}
We enrich the dataset by adding two types of sentiment scores. Firstly, we calculate a sentiment score using the spaCy library\footnote{\url{https://spacy.io/}} across all knowledge items selected for that dialogue. Secondly, we calculate a sentiment score specifically for the summary part of the ground truth responses. Our intuition is that the sentiment of the summary should align with the sentiment of the knowledge items used to generate the response. Our belief is supported by positive correlations observed in all data splits (train=0.41, val=0.53, test=0.54) and overall (all=0.45).

\begin{figure}[!h]
    \centering
    \includegraphics[width=.45\textwidth]{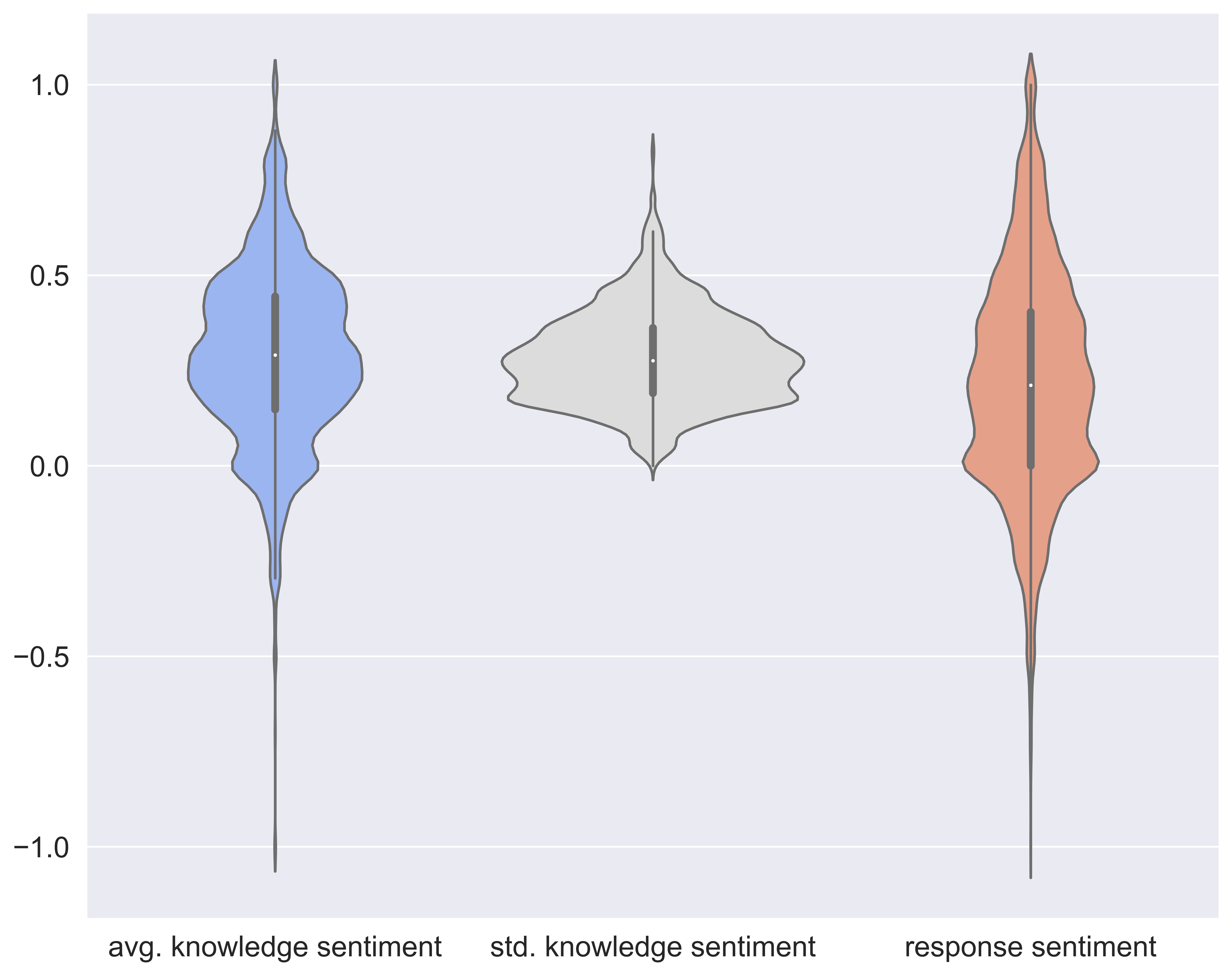}
    \caption{Sentiment distributions across dialogues for knowledge items (left) and responses (right). For knowledge items, we also show the distribution of standard deviation sentiment scores inside dialogues (centre).}
    \label{fig:distribution_sentiment}
\end{figure}

Furthermore, we examined the data distributions of these new features, as depicted in Figure~\ref{fig:distribution_sentiment}.
The average sentiment of the knowledge items and the sentiment of the responses have similar distributions. They both have positive medians, positive interquartile ranges, and a long tail extending into the negative range, indicating that the sentiment of the system's responses tends to be neutral to positive.
The average sentiment of the knowledge items is slightly higher than the sentiment of the responses.
The standard deviation of the spread of the knowledge sentiment scores inside dialogues is around 0.25, indicating that most knowledge items are not highly polarised, and their sentiment values are relatively close to each other.

\begin{table*}[!ht]
    \centering
    \scalebox{0.93}{
    \begin{tabular}{@{}l|cc|ccccc@{}}
        \toprule
        \textbf{Act} & \textbf{Freq.} & \textbf{Res. Len.}& \textbf{BLEU} & \textbf{METEOR} & \textbf{ROUGE-1} & \textbf{ROUGE-2} & \textbf{ROUGE-L}  \\
        \midrule
        yes\_no\_question & 1078 & 131.85 & 0.097 & 0.187 & 0.368 & 0.151 & 0.292 \\
        open\_question\_opinion & 288 & 133.52 & 0.107 & 0.189 & 0.379 & 0.163 & \textbf{0.301} \\
        open\_question\_factual & 203 & 134.51  & 0.094 & 0.179 & 0.358 & 0.148 & 0.285 \\
        statement & 188 & 137.44 & 0.081 & 0.170 & 0.345 & 0.131 & 0.260 \\
        command & 184 & 132.82 & 0.087 & 0.182 & 0.355 & 0.141 & 0.280 \\
        opinion & 166 & 139.07 & 0.089 & 0.183 & 0.366 & 0.143 & 0.280 \\
        pos\_answer & 14 & 135.78 & 0.053 & 0.151 & 0.312 & 0.108 & 0.230 \\
        complaint & 5 & 138.20 & \textbf{0.110} & 0.160 & 0.333 & 0.122 & 0.277 \\
        comment & 1 & 181.00 & 0.038 & 0.084 & 0.196 & 0.081 & 0.117 \\
        neg\_answer & 1 & \textbf{104.00} & 0.069 & \textbf{0.319} & \textbf{0.611} & \textbf{0.294} & 0.5 \\
        nonsense & 1 & 189.00 & 0.014 & 0.089 & 0.184 & 0.0 & 0.123 \\
        \bottomrule
    \end{tabular}}
    \caption{Data statistics (frequency and average system response length) and baseline performance per dialogue act.}
    \label{table:dialogue_act}
\end{table*}

\subsection{Dialogue management analysis}
\paragraph{Dialogue acts} We generate a dialogue act tag MIDAS \cite{midas-2019} for each of the user's last utterances in the dialogue context and explore the baseline performance per act type (Table \ref{table:dialogue_act}). Our findings reveal that questions are quite frequent, as users often rely on them to interact with the artificial assistant in task-oriented dialogues.  It is worth noting that \textit{yes/no questions} are the most frequent type and typically elicit short system responses, consisting of simple answers like "yes" or "no." Furthermore, we observe that some questions seek opinions, which may require subjective knowledge derived from reviews for an appropriate response, in contrast to factual questions that can potentially be answered using FAQs.

\paragraph{Optional questions}\label{sec:questions}We examined how frequently questions appear in the system's responses. Our findings indicate that 34.94\% of the responses contain a question at the end.

\begin{figure}[!h]
    \centering
    \includegraphics[width=.45\textwidth]{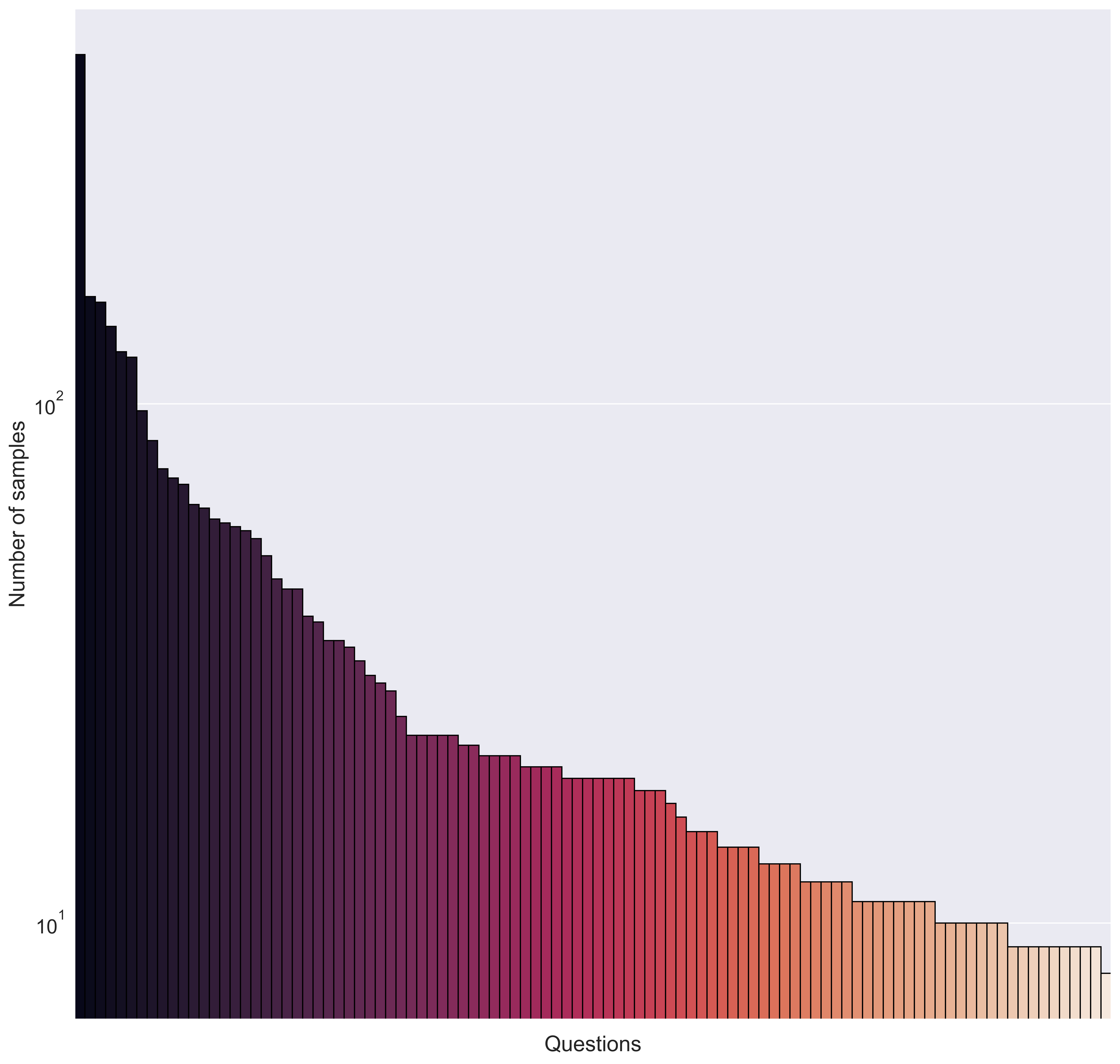}
    \caption{Question frequency distribution across all data splits. Only the first 100 elements are plotted.}
    \label{fig:distribution_questions}
\end{figure}

When analysing the questions, we observe that the data distribution exhibits a single peak and a long tail (Figure~\ref{fig:distribution_questions}). Out of the 2,522 unique questions, 79\% (n=1,994) appear only once. The most commonly occurring question is \textit{"Would you like to know more about them?}  accounting for 19\% of the occurrences. Moreover, we notice that the most frequent questions are generic. The top five questions, which make up 42\% of the data, include: \textit{"Do you have any other questions?"}, \textit{"Is there anything else I can help you with?"}, \textit{"Would you like to make a reservation?"}, and \textit{"Is there anything else you'd like to know about them?"}.

To evaluate the impact of optional questions, we conducted an ablation study where we remove the questions at the end of the responses (Table~\ref{table:questions}). This leads to slightly lower scores. Additionally, we experiment with adding the most frequent question (MFQ) to the responses that did not already contain a question. This improves the performance for longer responses.

\begin{table*}[!h]
    \centering
    \scalebox{0.93}{
    \begin{tabular}{@{}lccccc@{}}
        \toprule
        \textbf{Approach} & \textbf{BLEU} & \textbf{METEOR} & \textbf{ROUGE-1} & \textbf{ROUGE-2} & \textbf{ROUGE-L}  \\
        \midrule
         Baseline & \textbf{0.102} & 0.179 & \textbf{0.365} & 0.149 & \textbf{0.287} \\
         Strip questions & 0.083 & 0.165 & 0.359 & 0.148 & 0.284 \\
         Append MFQ & 0.094 & \textbf{0.181} & 0.363 & \textbf{0.150} & 0.286 \\
        \bottomrule
    \end{tabular}}
    \caption{Results of investigation regarding questions on the validation set.}
    \label{table:questions}
\end{table*}

\section{Method}
\subsection{Sub-task 2: Knowledge Selection}
Given the limited scope of reviews, traveller types and domains, we expand the knowledge base with few-shot data augmentation. Utilising ChatGPT, we synthesise artificial data, supplementing the MultiWoz 2.1 dataset's augmented version~\cite{eric-etal-2020-multiwoz}. This process involves creating new reviews for existing items within existing domains, in addition to the introduction of three new domains from the DSTC10 data set, all enriched with subjective knowledge.  In total, our additional data contains 1,006 extra reviews. We combine these with the existing data for use in both the Knowledge Seeking Turn Detection and Knowledge Selection tasks. For a data example see \ref{knowledge_example}.

\paragraph{Extension of item reviews}
The original knowledge data had 143 entities and 1,430 reviews (8,013 sentences) in the \textit{hotel} and \textit{restaurant} domains. We added 715 reviews 
to this by adding five different traveller types and producing five sentences for each type. This led to 25 extra sentences per entity,  for a total of 2,145 reviews and 11,609 sentences. We significantly expanded and diversified the present traveller types. The original data had only 6 types, we increased this to 26, though some like \textit{"Group of friends"} and \textit{"Friends getaway"} could be merged into the same type. For a complete overview see Table~\ref{tab:traveler_types}. The average sentence length for the original reviews is 73.73 compared to 76.93 for the added reviews. The prompt to generate these reviews can be found in Prompt \ref{prompt:newitems}

\paragraph{Extension to new domains}
Inspired by \citet{thulkeDSTC2021}, we expand our knowledge base by adding reviews from three different domains: \textit{attraction, taxi,} and \textit{train}.

\begin{theo}[extension of new reviews]{prompt:newitems}
Please provide new reviews for the \textbf{\{entity\_name\}} but for five different traveler\_type. \\ Continue the counting of the reviews and make sure the new reviews are in a dict format like this: \\
"<id>": \{\{"traveler\_type": "<traveler\_type>", "sentences": \{\{"<id>": "<review>", "<id>": "<review>", "<id>": "<review>", "<id>": "<review>"\}\}\}\}, \\

\noindent These are the existing reviews: \textbf{\{reviews\_all\}}. \\

\noindent Take this start and continue. Use double quotes to comply with json format. \\

\noindent "10": \{\{"traveler\_type":
\end{theo}

We use the validation data from DSTC10\footnote{Data downloaded using the following script: \href{https://github.com/dthulke/dstc10-track2/blob/main/code/i6_dstc10/datasets/data_augmentation.py}{https://tinyurl.com/j8pvkcnx}} which contains entities and respective FAQs from these different domains. The data did not contain subjective reviews for any of the domains. Hence, we generate three reviews for each entity by prompting ChatGPT with their entity name and FAQs.
Due to the limited context width of the ChatGPT API, we restrict the reviews to two sentences.

We generate the reviews with Prompt \ref{prompt:newdomains}, which resulted in $291$ reviews for the \textit{attraction} domain.
\begin{theo}[extension to new domains]{prompt:newdomains}
Given this example: \\ \\ \{
    \textbf{<entity\_id>}: \{
        "name": \textbf{<entity\_name>},
        "faqs": \{
            \textbf{<faq\_id>}: \{
                "question": \textbf{<question>},
                "answer": \textbf{<answer>}
            \}
        \}
    \}
\},\\ \\
can you generate three more reviews, not more than 2 sentences, as: traveler type: review?
\end{theo}

Although we prepared the data for the \textit{taxi} and \textit{train} reviews, we excluded these two domains as there were no transportation-related domains present in the released test set.

\subsection{Sub-task 3: Response Generation}
\subsubsection{Model exploration}
\label{sec:model_exploration}

\begin{table*}[!h]
    \centering
    \begin{tabular}{@{}lccccc@{}}
    \toprule
        \textbf{Model name} & \textbf{BLEU} & \textbf{METEOR} & \textbf{ROUGE-1} & \textbf{ROUGE-2} & \textbf{ROUGE-L} \\
    \midrule
        bart-large-cnn-samsum   & 0.095 & \textbf{0.184} & \textbf{0.363} & 0.144 & 0.278 \\
        bart-base               & 0.096 & 0.173 & 0.347 & 0.139 & 0.271 \\
        flan-t5-base            & 0.096 & 0.173 & 0.345 & 0.140 & 0.274 \\
        flan-t5-large           & \textbf{0.104} & 0.179 & 0.360 & \textbf{0.147} & \textbf{0.281} \\
        flan-t5-small           & 0.098 & 0.164 & 0.337 & 0.136 & 0.272 \\
    \bottomrule
    \end{tabular}
    \caption{Models explored, shown performance is on the test set. }
    \label{tab:models}
\end{table*}

The generation baseline provided by the organisers uses BART~\cite{lewis-etal-2020-bart}, specifically bart-base\footnote{\url{https://huggingface.co/facebook/bart-base}}, fine-tuned for ten epochs using Adam with a learning rate of 3e-5 and epsilon of 1e-8.

Instead of sticking to the base BART model, we opted to browse publicly available trained models on different tasks and investigate their performance after fine-tuning on the DSTC data. This approach is inspired by \emph{model recycling} \citep{choshen2022start}, which suggests that there exist models that are generally better at being adapted to perform domain-specific tasks.
We browsed the \texttt{HuggingFace Hub}\footnote{\url{https://huggingface.co/models}} for models with a better fit (in the task they were trained for or in the data they used for pre-training). Ultimately, we retrained the baseline architecture, as well as Flan-T5~\cite{chung2022scaling} in its small, base and large variants, and a trained BART model, trained on the SAMSum dataset~\cite{gliwa2019samsum}. See Table~\ref{tab:models} for the results. For training the larger architectures, including flan-t5-large, bart-large and flan-t5-base, we modified the optimisation parameters (learning rate 4e-05, a warm-up ratio of 0.2). We found that using a larger model has small benefits over the original baseline model in most evaluation metrics. We did not find that training on intermediate tasks that are related to our objective improved performance substantially.

\subsubsection{Prompting}
We used OpenAI's APIs\footnote{\url{https://platform.openai.com/docs/api-reference}} to generate system responses given the baseline output for sub-task 1 and 2. To ensure consistent results, we set the temperature value to 0 for all configurations.

\begin{table*}[!h]
    \centering
    \begin{tabular}{@{}lccccc@{}}
    \toprule
        \textbf{Model name} & \textbf{BLEU} & \textbf{METEOR} & \textbf{ROUGE-1} & \textbf{ROUGE-2} & \textbf{ROUGE-L} \\
    \midrule
        ChatGPT                               & 0.041 & 0.147 & 0.280 & 0.085 & 0.207 \\
        GPT-3                             & 0.051 & 0.137 & 0.302 & 0.105 & 0.237 \\
        ChatGPT + CoT              & 0.050 & 0.155 & 0.305 & 0.099 & 0.228 \\
        ChatGPT + CoT + 3 examples & \textbf{0.069} & \textbf{0.161} & \textbf{0.325} & \textbf{0.117} & \textbf{0.243} \\
        Waterfall                            & 0.056 & 0.152 & 0.312 & 0.100 & 0.236 \\
        Waterfall + 3 examples               & 0.057 & 0.160 & 0.311 & 0.103 & 0.225 \\
    \bottomrule
    \end{tabular}
    \caption{Prompting settings explored, shown performance is on the validation set. }
    \label{table:prompting}
\end{table*}

\paragraph{Model comparison}
To compare GPT-3 and ChatGPT, we conducted experiments using simple prompts while adhering to the respective API requirements. The prompts consisted of two components: a) the dialogue history and b) the selected subjective knowledge. The subjective knowledge was formatted as a list, explicitly specifying whether each item was a \textbf{Review} or an \textbf{FAQ}.

\begin{verbatim}
    {knowledge_type}: {text}
\end{verbatim}

\noindent The initial prompt for GPT-3 is shown in Prompt \ref{prompt:initialGPT3}

\begin{theo}[GPT-3, initial prompt]{prompt:initialGPT3}
DIALOGUE:\\
    \textbf{\{dialogue\_context\}} \\

\noindent KNOWLEDGE:\\
    \textbf{\{knowledge\}} \\

\noindent RESPONSE:
\end{theo}

ChatGPT requires API requests to be given as a series of messages with a "role" that can have three values: \textit{System}, \textit{User} or \textit{Assistant}. As they explain in their documentation, \textit{System} messages can be used to give instructions to the model, while \textit{User} and \textit{Assistant} messages can be used to carry on a dialogue with the model. As such, we use the \textit{System} messages to give the subjective knowledge to the model and then send each of the dialogue history messages as either \textit{User} or \textit{Assistant} messages. The initial prompt for ChatGPT is shown in Prompt \ref{prompt:initialChatGPT}.

\begin{theo}[ChatGPT, initial prompt]{prompt:initialChatGPT}
  \{"role": "system", \\
    "content": "You are a helpful assistant with access to the following:\\
                    \textbf{\{knowledge\}}" \}, \\
  \{"role": "user", \\
    "content": "\textbf{\{dialogue\_context\}}" \}, \\
  \{"role": "assistant", \\
    "content": "\textbf{\{dialogue\_context\}}" \}, \\
  \{"role": "user", \\
    "content": "\textbf{\{dialogue\_context\}}" \} \\
\end{theo}

Although ChatGPT is optimised for dialogue and we expected it to perform better, it scored lower than GPT-3 in the automatic metric evaluation, as seen in Table~\ref{table:prompting}. Upon further examination of the generated responses, we observed that ChatGPT tends to produce much longer responses. In fact, 367 API requests were truncated due to their length. On the other hand, GPT-3 generates more concise responses compared to both the baseline and the ground truth (Table~\ref{table:length-stats}). This difference in response length has a negative impact on most natural language generation metrics, as they tend to penalise longer and wordier responses.

\begin{table}[!h]
    \centering
    \scalebox{0.9}{
    \begin{tabular}{@{}lcccc@{}}
        \toprule
        \multirow{2}{*}{\textbf{Approach}} & \multicolumn{2}{c}{\textbf{\#Sentences}} & \multicolumn{2}{c}{\textbf{\#Characters}}  \\
        ~ & \textbf{Avg.} & \textbf{Std.} & \textbf{Avg.} & \textbf{Std.} \\
        \midrule
         GPT-3                  & 1.55 & 0.77 & 120.04 & 70.16 \\
         Ground truth (val)     & 1.69 & 0.66 & 133.55 & 32.05 \\
         Ground truth (test)    & 1.70 & 0.66 & 135.55 & 32.12 \\
         Ground truth (train)   & 1.71 & 0.65 & 136.61 & 33.49 \\
         Baseline (val)         & 1.91 & 0.53 & 129.66 & 29.07 \\
         ChatGPT                & 2.09 & 0.79 & 182.62 & 74.61 \\
        \bottomrule
    \end{tabular}}
    \caption{Length statistics on responses, ordered by ascending average length}
    \label{table:length-stats}
\end{table}

\paragraph{Chain Of Thought}
We focus on optimising the prompts for the best performance of a single model, specifically ChatGPT. To achieve this, we adopt recommended techniques from the OpenAI Cookbook\footnote{\url{https://github.com/openai/openai-cookbook/blob/main/techniques_to_improve_reliability.md}} for prompting ChatGPT, such as Chain of Thought (CoT)~\cite{wei2022chain}. For this, we divide the task into two sub-tasks: summarisation and dialogue management (See Section~\ref{sec:data_analysis}). The resulting prompt is shown in Prompt \ref{prompt:chainthought}.

To accommodate CoT answers, which tend to be longer, we double the value of the "max\_tokens" parameter. This modification leads to improved performance (refer to Table~\ref{table:prompting}); however, it also increases the number of truncated API responses to 1,110.

\begin{theo}[ChatGPT, Chain Of Thought]{prompt:chainthought}
You are assisting a user. Create a response for the user, using the following procedure: \\
\indent \textit{(1) First, summarise the available knowledge into a couple sentences.} \\
\indent \textit{(2) Then, create a short follow-up question given the dialogue history.} \\
\indent \textit{(3) Create the final response to the user as <summary><follow-up>} \\

\noindent Knowledge: \\
\indent \textbf{\{knowledge\} }\\

\noindent Dialogue history: \\
\indent \textbf{\{dialogue\_context\}} \\

\noindent Solution:\\
\indent \textit{(1) summary}:
\end{theo}

\paragraph{Few-shot learning}
In order to further enhance the prompt, we incorporate few-shot learning~\cite{brown2020language}. After initial exploration, we select three examples that pose the most difficulty. To identify these examples, we analyse instances with the lowest performance and examine their data trends. We observe that ChatGPT struggles with dialogues that involve a high number of items and have a predominantly negative sentiment. Consequently, we select dialogues from the training dataset that meet the following criteria:

\begin{enumerate}
    \item Filter by the number of knowledge items that are "higher than usual", between [mean+std, mean+2std]
    \item  Filter by the average sentiment that is "lower than usual", between [mean-2std, mean-std]
    \item Since around one-third of responses contain a question, we sample accordingly
\end{enumerate}

By applying these selection criteria, we focus on dialogues that require better summarisation due to the increased number of knowledge items with slightly negative sentiment, and a mix of dialogues with and without a question. This setting yields the best performance and reduces the number of truncated API responses to 380.

\paragraph{Waterfall prompting}
Additionally, we explore combining prompts and models by utilising the output of one prompted model as part of the input prompt for a subsequent model, in a waterfall manner. We use GPT-3 for summarisation since it behaves better at this part of the task~\cite{bhaskar2023prompted}. We also separate the FAQs, containing factual information, from the reviews, which have opinionated information. The prompt we use for this task is shown in Prompt \ref{prompt:summarisation}.

\begin{theo}[GPT-3, Summarisation]{prompt:summarisation}
Summarize the following into one or two sentences max: \\

\noindent FAQs: \\
\indent \textbf{\{faqs\}} \\

\noindent Reviews: \\
\indent \textbf{\{reviews\}} \\
\end{theo}

We then use ChatGPT to generate the system response in dialogue, using the output of GPT-3 as part of the prompt. The final prompt is shown in Prompt \ref{prompt:final}. Interestingly, we observed a decline in performance when using this waterfall approach compared to a single model setting.

\begin{theo}[ChatGPT, Final prompt]{prompt:final}
You are assisting a user. Create a response for the user, using the following procedure: \\
\textit{\indent (1) First, summarise the available knowledge into a couple sentences. \\
\indent (2) Then, create a short follow-up question given the dialogue history. \\
\indent (3) Create a final brief response to the user as <summary><follow-up> \\}

\noindent Knowledge: \\
\indent \textbf{\{knowledge\}} \\

\noindent Dialogue history: \\
\indent \textbf{\{dialogue\_context\}} \\

\noindent Solution:\\
\textit{\indent (1) summary:} \\
\indent \indent \textbf{\{GPT-3 summary\}} \\
\textit{\indent (2) follow-up: }
\end{theo}

\section{Results}
Table \ref{table:results} shows the submitted approaches' results for all three sub-tasks. We ranked as the 8th best team; hence we did not get official human evaluation results on our submissions. Two of our approaches perform better than the baseline on automatic evaluation metrics, and our identifier in the official results is Team 10.

In the sub-task of \textbf{Knowledge Seeking Turn Detection}, our added knowledge induced a slight decrement in performance, with precision declining from 0.998 to 0.995 and F1 score from 0.998 to 0.997, albeit recall experienced a marginal increase (refer to Table \ref{table:results} for further details).

Conversely, \textbf{Knowledge Selection} saw an improvement with the augmentation, resulting in a small uplift across all metrics: precision from 0.790 to 0.796, recall from 0.788 to 0.794, F1 from 0.789 to 0.795, and Exact Match from 0.91 to 0.418.
\begin{table*}[ht!]
\scriptsize{
    \centering
    \begin{tabular}{@{}lccc|cccc|ccccc|c@{}}
        \toprule
        \textbf{Approach} & \multicolumn{3}{c}{\textbf{sub-task 1}} & \multicolumn{4}{c}{\textbf{sub-task 2}} & \multicolumn{5}{c}{\textbf{sub-task 3}} & \textbf{Total}  \\
        \midrule
         & P & R & F1 & P & R & F1 & EM & BLEU & METEOR & ROUGE-1 & ROUGE-2 & ROUGE-L & MRR  \\
        \midrule
        Baseline & 0.998 & 0.998 & 0.998 & 0.790 & 0.788 & 0.789 & 0.391 & 0.100 & 0.175 & 0.352 & 0.143 & 0.275  & 0.039 \\
        \midrule
        CLTeamL-0 & 0.995 & 0.999 & 0.997 & 0.796 & 0.794 & 0.795 & 0.418 & 0.104 & 0.179 & 0.360 & 0.147 & 0.281 & 0.068 \\
        CLTeamL-1 & 0.995 & 0.999 & 0.997 & 0.796 & 0.794 & 0.795 & 0.418 & 0.094 & 0.181 & 0.356 & 0.146 & 0.279 & 0.045 \\
        CLTeamL-2 & 0.996 & 0.998 & 0.997 & 0.772 & 0.818 & 0.794 & 0.423 & 0.060 & 0.121 & 0.241 & 0.087 & 0.186 & 0.025 \\
        \midrule
        CLTeamL-2b & 0.996 & 0.998 & 0.997 & 0.772 & 0.818 & 0.794 & 0.423 & 0.068 & 0.157 & 0.311 & 0.112 & 0.232 & NA \\
        \bottomrule
    \end{tabular}
    \caption{Results over the test set. The final ranking is determined by the mean reciprocal rank (MRR) of all scores.}
    \label{table:results}}
\end{table*}

We submitted three different approaches for the \textbf{Response Generation} task. (1) Our first submission is based on the model exploration mentioned in \ref{sec:model_exploration}. After fine-tuning flan-t5-large this approach outperforms the baseline on all metrics, leading to a mean reciprocal rank (MRR) of 0.068 compared to 0.039 of the baseline. On BLEU this approach ranked 7th out of 49 submissions.
(2) Based on our data analysis (\ref{sec:questions}), we extended this approach in post-processing with the most frequently appearing question from the training set appended to the end of the review summaries. This approach, although underperforming relative to our first except in terms of METEOR, still surpassed the baseline on all metrics, apart from BLEU, with an MRR of 0.045.
(3) Despite lower performance relative to the baseline in quantitative metrics, our waterfall prompting approach involving a combination of ChatGPT and GPT-3 gave qualitatively promising results. Presented with non-matching knowledge items it did not generate a summary, but flagged them as not relevant to the questions at hand. A full human evaluation would be an interesting step to investigate the perceived accuracy and appropriateness of such responses.

\subsection{Submissions}
\paragraph{CLTeamL-0} Baseline with augmented data for Knowledge-seeking Turn Detection and Knowledge Selection, flan-t5-large for Response Generation 

\paragraph{CLTeamL-1} Baseline with augmented data for Knowledge-seeking Turn Detection and Knowledge Selection, flan-t5-large for Response Generation. Additionally, we do post-processing on the output, adding the most frequent question from the training set as a follow-up question to each review summary. 

\paragraph{CLTeamL-2} Baseline for Knowledge-seeking Turn Detection and Knowledge Selection. For the Response Generation, we prompted ChatGPT with Chain Of Thought instructions, to first generate a summary of the knowledge and then append a follow-up question given the dialogue history. We provided three examples, selected from the most challenging cases in the training data according to the data trends on the validation set. Unfortunately, due to server problems, the data samples after 4500, are lacking a review summary. A full run (see Approach 2b in Table \ref{table:results}) after submission showed slight improvements across all metrics but still did not outperform the baseline or our other approaches. 

\section{Conclusion}
In conclusion, our work on the Task-Oriented Conversational Modeling with Subjective Knowledge task produced three key contributions: Firstly, our detailed data analysis could serve as a basis for future dataset scoping within this domain. Secondly, we significantly expanded the knowledge dataset size utilising few-shot data augmentation. Lastly, our most successful model was a fusion of the baseline with augmented data and flan-t5-large. Our waterfall prompting approach incorporating a blend of Large Language Models demonstrated lower metrics compared to the baseline, but upon a qualitative assessment, the results were deemed satisfactory, albeit the absence of official human evaluation impedes a definitive judgement.

A preliminary qualitative analysis shows that ChatGPT can spot mistakes during knowledge selection. Thus, future work could explore incorporating an initial step where ChatGPT is employed to evaluate the relevance of the selected knowledge items. Additionally, we plan to experiment with summarising positive and negative reviews separately before creating a consolidated summary. This method may enhance performance when dealing with polarised reviews.

\section*{Limitations}
The APIs provided by OpenAI come with particular length parameters which pose certain constraints in terms of flexibility and generalisability to larger datasets and longer sequences which are prevalent in knowledge-grounded dialogue.

There's also a financial aspect to consider, as the use of these APIs involves a monetary cost. We chose to proceed with ChatGPT instead of GPT-3 for parts of our research, even though preliminary explorations indicated higher performance by the latter, due to the additional cost associated with the use of GPT-3. Hence, there might be potential performance enhancements our approaches were not able to capture due to this cost-based decision.

\section*{Ethics Statement}
We artificially increased the size of our knowledge data set with few-shot data augmentation to cover a broader range of traveller types and domains. The synthetic reviews were created using GPT-3 and ChatGPT. There exists a potential for the introduction of biases in this newly generated data due to inherent model biases that might have been propagated into the reviews. This is especially concerning for commercial LLMs, like the ones used, since their training data is not made public.

\section*{Acknowledgements}
This research was funded by the Vrije Universiteit Amsterdam and the Netherlands Organisation for Scientific Research (NWO) through the \textit{Hybrid Intelligence Centre} via the Zwaartekracht grant (024.004.022), and the \textit{Spinoza} grant (SPI 63-260) awarded to Piek Vossen.

\bibliography{anthology,custom}
\bibliographystyle{acl_natbib}

\onecolumn
\appendix

\section{Appendix}
\label{sec:appendix}

\begin{lstlisting}[language=json,basicstyle=\scriptsize,columns=flexible]
{
  "hotel": {
    "0": {
      "name": "A AND B GUEST HOUSE",
      "reviews": {
        "0": {
          "traveler_type": "Solo travelers",
          "sentences": {
            "0": "I was really happy with my recent stay at A and B Guest House.",
            "1": "I stayed on my own, and I'm a smoker, so I was super happy that there was a designated area especially for smokers.",
            "2": "I also thought that my room was very spacious, and I was pleased with the breakfast options that were available."
          }
        },
        "1": {
          "traveler_type": "Couples",
          "sentences": {
            "0": "My husband was pleased to be able to park on site for free.",
            "1": "We thought it was a bit noisy at A and B especially because it was just us and we had looked forward to quiet."
          }
        },
        "2": {
          "traveler_type": "Business travelers",
          "sentences": {
            "0": "I wasn't thrilled the with lack of conveniences at A and B Guest House on my recent business trip.",
            "1": "No luggage storage service was inconvenient, and I was surprised there was no safety deposit box in a room.",
            "2": "I do appreciate the good breakfast options and the comfort of the bed, though.",
            "3": "So overall, I would probably stay there again."
          }
        },
        "3": {
          "traveler_type": "Solo travelers",
          "sentences": {
            "0": "I stayed at this guesthouse by myself.",
            "1": "I was happy with the laundry facilities that were on site.",
            "2": "The free available Internet provided great speeds and a strong connection.",
            "3": "I was treated to a lovely filling breakfast in the morning.",
            "4": "However, when I went to use the bathroom I found that the bathroom was not clean.",
            "5": "Also, when I went to put away my luggage I found that there were no provided spaces in which to do so."
          }
        },
        "4": {
          "traveler_type": "Solo travelers",
          "sentences": {
            "0": "Not long ago, I stayed at A and B Guest House by myself.",
            "1": "The staff was really friendly when I checked in, and I was thrilled with how comfy the bed was.",
            "2": "I got a great night of sleep!",
            "3": "But I was sad that there was no elevator so I had to use the stairs, and overall I think the room was overpriced."
          }
        },
        "5": {
          "traveler_type": "Solo travelers",
          "sentences": {
            "0": "I stayed here by myself, and it was nice to be able to wash my clothes at the laundry.",
            "1": "Looking out of the window was nice, as the view was unique.",
            "2": "The location makes it a delight to get around on foot, and see things.",
            "3": "If only the room was a bit cleaner, it would have been perfect."
          }
        },
        "6": {
          "traveler_type": "Solo travelers",
          "sentences": {
            "0": "The A and B Guest House is a great place to stay.",
            "1": "It is a beautiful place in a good location.",
            "2": "Very quiet.",
            "3": "The staff went out of their way to make sure that our stay was outstanding.",
            "4": "The clerk even sent a hand drawn card to us when he found it was our anniversary.",
            "5": "I don't remember when I've slept in a bed more comfortable than the one here.",
            "6": "Fair warning if you travel with valuables; there is no safety box in the room.",
            "7": "Be careful."
          }
        },
        "7": {
          "traveler_type": "Families",
          "sentences": {
            "0": "My family and I decided to stay here while we were in town.",
            "1": "It was kid friendly, so my kids weren't treated like garbage.",
            "2": "It was also in a good location with access to most important things, and parking was completely free.",
            "3": "The staff, on the other hand, were really unprofessional and laid back.",
            "4": "They were not professional at all."
          }
        },
        "8": {
          "traveler_type": "Solo travelers",
          "sentences": {
            "0": "The room was very large and spacious.",
            "1": "Made it very easy to settle in and relax after a long trip, and even more relaxing with the excellent view out of my window!",
            "2": "The room included free Wi-Fi which was very much appreciated because its quite uncommon to get free internet service these days!",
            "3": "My only complaint was that there was no concierge service and I had to take care of this myself - a bit annoying but overall a great stay."
          }
        },
        "9": {
          "traveler_type": "Couples",
          "sentences": {
            "0": "The A and B could be a problem if you need an accessible room you may need to book elsewhere.",
            "1": "The room was not very clean but it was large.",
            "2": "My boyfriend and I both thought the bed had a good firmness for our liking.",
            "3": "There is a great view especially at sunset."
          }
        "10": {
          "traveler_type": "Family travelers",
          "sentences": {
            "0": "We had a great time staying at A and B Guest House with our kids.",
            "1": "The staff was very accommodating and provided us with a baby crib and high chair upon request.",
            "2": "The breakfast options were great and the location was convenient for our family activities.",
            "3": "The only downside was that the room was a bit cramped for our family of four."
          }
        },
        "11": {
          "traveler_type": "Group travelers",
          "sentences": {
            "0": "Our group of friends stayed at A and B Guest House and we had a great time.",
            "1": "The staff was friendly and helpful, and the location was perfect for exploring the city.",
            "2": "The rooms were clean and comfortable, and the breakfast was delicious.",
            "3": "The only issue we had was that the Wi-Fi was a bit spotty in some areas of the guest house."
          }
        },
        "12": {
          "traveler_type": "Budget travelers",
          "sentences": {
            "0": "A and B Guest House was a great choice for our budget-friendly trip.",
            "1": "The room was clean and comfortable, and the breakfast options were good.",
            "2": "The location was also convenient for exploring the city on foot.",
            "3": "The only downside was that the room was a bit small and there was no elevator."
          }
        },
        "13": {
          "traveler_type": "Luxury travelers",
          "sentences": {
            "0": "As a luxury traveler, I was disappointed with my stay at A and B Guest House.",
            "1": "The room was small and lacked the amenities I expect from a luxury hotel.",
            "2": "The staff was friendly but not particularly attentive to my needs.",
            "3": "The only positive was the location, which was convenient for exploring the city."
          }
        },
        "14": {
          "traveler_type": "Pet owners",
          "sentences": {
            "0": "We were happy to find a pet-friendly option in A and B Guest House.",
            "1": "The staff was accommodating and provided us with a pet bed and bowls upon request.",
            "2": "The location was also convenient for walking our dog.",
            "3": "The only downside was that there was no designated pet area for our dog to use the bathroom."
          }
        }
        }
      },
      "faqs": {
        "0": {
          "question": "Are children welcomed at this location?",
          "answer": "Yes, you can stay with children at A and B Guest House."
        },
        "1": {
          "question": "Can I bring my pet to A and B Guest House?",
          "answer": "No, pets are not allowed at this property."
        },
        "2": {
          "question": "Do you have onsite parking for your guests?",
          "answer": "There is onsite parking at A and B Guest House but it costs extra."
        },
        "3": {
          "question": "What time is check-in there?",
          "answer": "Check-in time is from 3:30pm - 9:00pm."
        },
        "4": {
          "question": "Is smoking allowed on the property?",
          "answer": "There are designated smoking areas throughout"
        },
        "5": {
          "question": "What languages are spoken?",
          "answer": "English, Italian, Lithuanian, Portuguese, and Russian are spoken here."
        },
        "6": {
          "question": "Should I make a reservation for parking?",
          "answer": "You need to make a reservation at A and B Guest House for parking."
        },
        "7": {
          "question": "Are children allowed to check in here?",
          "answer": "An individual has to be 18 and over to check in at A and B Guest House."
        },
        "8": {
          "question": "what time do I check out?",
          "answer": "Check out times range from 7:30 AM to 10:00 AM."
        },
        "9": {
          "question": "Can my small dog stay with me?",
          "answer": "Pets are not allowed at the A and B Guest House."
        }
      }
    },
\end{lstlisting}
\paragraph{Knowledge data example:}\label{knowledge_example} Slightly shortened example from the knowledge.json. Reviews 0-9 are original, and reviews 10-14 are augmented.

\begin{table}[ht]
\centering
\begin{tabular}{@{}lrr@{}}
\toprule
\textbf{Traveler Type} & \textbf{Count Original Reviews} & \textbf{Count Augmented Reviews} \\
\midrule
Families & 320 & 95 \\
Solo travelers & 315 & 101 \\
Couples & 279 & 109 \\
Colleagues & 230 & 11 \\
Friends & 210 & 84 \\
Business travelers & 76 & 111 \\
Budget travelers & 0 & 49 \\
Luxury travelers & 0 & 39 \\
Families with children & 0 & 22 \\
Foodies & 0 & 19 \\
Vegetarians & 0 & 10 \\
Pet owners & 0 & 9 \\
Business & 0 & 6 \\
Groups & 0 & 6 \\
Backpackers & 0 & 5 \\
Friends getaway & 0 & 5 \\
Families with teenagers & 0 & 3 \\
Adventure travelers & 0 & 3 \\
Retirees & 0 & 3 \\
Group of friends & 0 & 2 \\
Families with young children & 0 & 2 \\
Honeymooners & 0 & 1 \\
Senior travelers & 0 & 1 \\
Romantic couples & 0 & 1 \\
Romantic travelers & 0 & 1 \\
Students & 0 & 1 \\
\bottomrule
\end{tabular}
\caption{Comparison of traveler types between the original reviews and the added reviews. Some extremely similar types were combined into one e.g. "Family with children" and "Family with kids".}
\label{tab:traveler_types}
\end{table}

\end{document}